\documentclass[journal]{IEEEtran} 
\IEEEoverridecommandlockouts
\usepackage{cite}
\usepackage{amsmath,amssymb,amsfonts}
\usepackage{algorithmic}
\usepackage{graphicx}
\usepackage{textcomp}
\usepackage{xcolor}
\usepackage{booktabs}
\usepackage{multicol}
\usepackage{ulem}
\usepackage{xcolor}
\usepackage{hyperref}
\def\BibTeX{{\rm B\kern-.05em{\sc i\kern-.025em b}\kern-.08em
    T\kern-.1667em\lower.7ex\hbox{E}\kern-.125emX}}
\begin{document}

\title{Data-Driven Radio Propagation Modeling using Graph Neural Networks\\
\thanks{Orange Labs}
}

\author{
    \IEEEauthorblockN{Adrien Bufort}
    , \IEEEauthorblockN{Laurent Lebocq}
    , \IEEEauthorblockN{Stéfan Cathabard}
    \\
    \IEEEauthorblockA{\textit{Orange Labs, Belfort, France}}\\
}

\maketitle

\begin{abstract}
Modeling radio propagation is essential for wireless network design and performance optimization. Traditional methods rely on physics models of radio propagation, which can be inaccurate or inflexible. In this work, we propose using graph neural networks to learn radio propagation behaviors directly from real-world network data. Our approach converts the radio propagation environment into a graph representation, with nodes corresponding to locations and edges representing spatial and ray-tracing relationships between locations. The graph is generated by converting images of the environment into a graph structure, with specific relationships between nodes. The model is trained on this graph representation, using sensor measurements as target data.

We demonstrate that the graph neural network, which learns to predict radio propagation directly from data, achieves competitive performance compared to traditional heuristic models. This data-driven approach outperforms classic numerical solvers in terms of both speed and accuracy. To the best of our knowledge, we are the first to apply graph neural networks to real-world radio propagation data to generate coverage maps, enabling generative models of signal propagation with point measurements only.
\end{abstract}


\section{Introduction}
Radio propagation is a crucial aspect of wireless communication and has been the subject of extensive research for decades. Accurate models of radio propagation are essential for the design and optimization of wireless networks. Recently, the field of machine learning has shown tremendous promise in a variety of applications, including the modeling of radio propagation. In this work, we will use deep learning (and specifically graph neural network \cite{b0}) to create radio propagation model that is able to generate radio coverage without any physical heuristic.

Currently "classic" numerical solver dominate the radio propagation simulator landscape but a lot of work have already been done to create machine learning based radio propagation model \cite{b2}. For example work have been done to distill classic solver into neural network using simulated data to calibrated neural network for indoor or outdoor environment (\cite{b1, b3}). Using real world data it is possible to create a radio propagation simulator from scratch that take into account the geographic topology (buildings, forest etc) (\cite{b7, b8}). Those work often forecast only point values which limit their ability to be fast to estimate a coverage map (due to the necessity of doing features engineering for each points).

In this work we will have several contributions :

- Create a \textbf{specific neural network architecture} that enable ray tracing behaviour to be incorporated smoothly into the neural network output. We will use graph neural network, a specific kind of neural network that can "pass" information from points to points according to a graph topology.

- \textbf{A training procedure} to learn to do radio propagation coverage from point measurements using neural masking to create models that are able to output full coverage of radio propagation signal but trained only on point measurements, which demonstrates its ability to generalize and generate coverage maps from limited data.

\section{Related Work}

\subsection{Machine learning and neural networks}

Machine learning is a sub-field of artificial intelligence that focuses on the development of algorithms that can learn patterns in data and make predictions based on that learning. Supervised learning is a type of machine learning in which the algorithms are trained on labeled data, and the goal is to learn a mapping from input variables to output variables. 

Supervised learning can be mathematically described as a function approximation problem. Given a set of input-output pairs, $(x_1, y_1), (x_2, y_2), ..., (x_n, y_n)$, the goal is to find a function $f(x)$ that maps input variables x to output variables y such that $f(x_i) \approx y_i$ for all $i = 1, 2, ..., n$.
The goal is to find the parameters of the function $f(x)$ that minimize the average loss over the entire dataset. The loss here is often the Mean Square Error (MSE) :  \begin{equation}
MSE = \frac{1}{n} \sum_{i=1}^{n} (f(x_i) - y_i)^2
\end{equation}
This process is known as training and the learned function $f(x)$ is then used for prediction on new, unseen data. There is a various set of learning algorithm, in this work we will choose artificial neural network as our main support to learn.

An artificial neural network (ANN) is a type of machine learning model. It consists of interconnected nodes, called artificial neurons, which are organized into layers. We refer the reader to the extensive litterature around neural network for a better understanding of its properties \cite{b9, b10}.

The ANN is a function $f(x;\theta)$ parameterized by weights $\theta$, the goal of training is to find the optimal weights $\theta^*$ that minimize the loss function $L(\theta)$:

\begin{equation}
\theta^* = \arg\min_{\theta} L(\theta)
\end{equation}

\subsection{Graph neural networks}

A graph neural network (GNN) is a type of neural network designed to process graph-structured data. In a graph, nodes represent entities and edges represent relationships between those entities. A GNN operates on the nodes of the graph and leverages the relationships between them to make predictions. Unlike traditional neural networks, which are designed for Euclidean structured data, GNNs can effectively handle non-Euclidean structured data, such as social networks, molecular graphs, and road networks. GNN have also been successfully applied to simulate physics \cite{b4, b5, b6} and this motivated our approach to use them.

Each node in a GNN has a feature representation, denoted as $h_i$, which summarizes the information of the node. The feature representations are updated iteratively based on the representations of its neighboring nodes, using a message-passing mechanism. Mathematically, the update can be described as two functions :

Message from node $j$ to node $i$:
\begin{equation}m_{j \rightarrow i} = MessageFunction(h_j, h_i, e_{j \rightarrow i})
\end{equation}

New representation of node $i$:
\begin{equation}h_i' = ReduceFunction({m_{j \rightarrow i} \mid j \in \mathcal{N}_i})
\end{equation}

In these equations, $i$ and $j$ represent nodes in the graph, $h$ represents the hidden state of a node, $e$ represents the edge between two nodes, and $m$ represents the message passed between two nodes. $\mathcal{N}_i$ represents the set of neighboring nodes of node $i$, and $\text{MessageFunction}$ and $\text{ReduceFunction}$ are functions that operate on the hidden states and edges to generate messages and reduce them to new node representations.

In this work, we employ a variant of the graph neural network called graphNetworks blocks (Message Passing GNN) \cite{b6}. We choose this neural architecture for two main reasons:

We wanted to leverage the invariance property of GNN, which takes into account the relations between positions.
We wanted to benefit from the topological flexibility of GNN, allowing us to create a customized graph to better account for ray tracing behavior.

\subsection{Masked output for training}

In a masked output training procedure, a portion of the output is masked, or hidden, during training, and the neural network must predict the masked values. This can be used to evaluate the performance of the neural network on previously unseen data, or to perform multi-task learning, where the neural network must perform multiple prediction tasks simultaneously.

Let $y$ be the ground truth output and $\hat{y}$ be the predicted output, with a portion of the output denoted as $m$, where $m_i=1$ indicates that the $i$-th component of the output is masked (which means we have the information about the data point) and $m_i=0$ indicates that it is not masked (we don't have any information). The loss function for masked output training is defined as:

\begin{equation}L(\theta) = \frac{1}{n} \sum_{i=1}^{n} m_i (y_i - \hat{y}_i(\theta))^2
\end{equation}

where $\theta$ are the parameters of the neural network, $n$ is the number of samples, and $(y_i, \hat{y}_i(\theta))$ is the $i$-th sample of the ground truth output and the predicted output, respectively. The loss function only penalizes the masked components of the output, allowing the neural network to make predictions for the unmasked components while learning to predict the masked components. The parameters $\theta$ are updated during training to minimize the loss function, using a optimization algorithm such as gradient descent or Adam.

We will use the masked output training procedure in our case because we only have partial measurement concerning the area we want to forecast.

GNN are particulary useful for doing this kind of semi-supervised learning and have been design to handle those partial target behaviour \cite{b16}.

\section{Data Collection and Preprocessing}

To create our radio propagation model, we utilized three main datasets: the measurement dataset, the geographic dataset, and the antenna dataset. The measurement dataset consists of actual field measurements of radio signal strength, obtained from various locations in France. The geographic dataset provides information about the surrounding environment, including building type, building height, ground height, and other relevant factors that may affect radio propagation. Finally, the antenna dataset contains information about the antenna configurations used in the measurements, such as azimuth, height, frequency, and antenna diagram.

Before we could use these datasets to train our machine learning model, we performed several preprocessing steps to ensure that the data was clean and appropriately formatted. In the following sections, we describe the details of each dataset and the preprocessing steps performed to prepare the data for machine learning.

\subsection{Overview of the dataset and its characteristics}

Here is a quick description of the different datasets used for the training and validation : 

- \textbf{An actual field measurement dataset} that contains 300M points of signal power measurement on all the territory of France in the year 2022. 

- \textbf{The geographic context} about the position of buildings near the antennas (buildings heights / type of building structure) and also the ground height. We limit ourself to a surface of 2kmx2km at a resolution of 5m (which create a 400x400 pixels images).

- \textbf{The antenna configuration} and antenna diagram (the azimut and tilt of the antenna, its height, its frequency, its antenna gains and its EIRP (Effective Isotropic Radiated Power)) of all the set of antennas in France.

\subsection{Dataset of measurements}

Thanks to the collection of data through the 'Orange et Moi' mobile application, we have access to 300 million data points of cell signal power measurements across France (figures \ref{francepoint} and \ref{paris}). These measurements were taken by Orange employees and are limited to 4G cells only. Each measurement point contains information such as the GPS coordinates, which are obtained directly from the mobile and have a high level of precision (with an estimated location error of less than 20 meters), making them critical for the model's performance.

Other information available in our dataset includes the ID of the connected cell (from the 20,000 sites and 200,000 cells in the dataset), the date of the measurement (limited to the year 2022 and provided in UTC), the actual signal power received from the connected cell (in dB, although the precision of this value is speculative and dependent on the mobile type and configuration), and the speed of the mobile/user. The latter is a strong indicator of the type of environment the mobile is in.

To get an idea of the dataset scale, here is an image of all the points of measurements in France and in the Paris city :
\begin{figure}[!htb]
\includegraphics[scale=0.24]{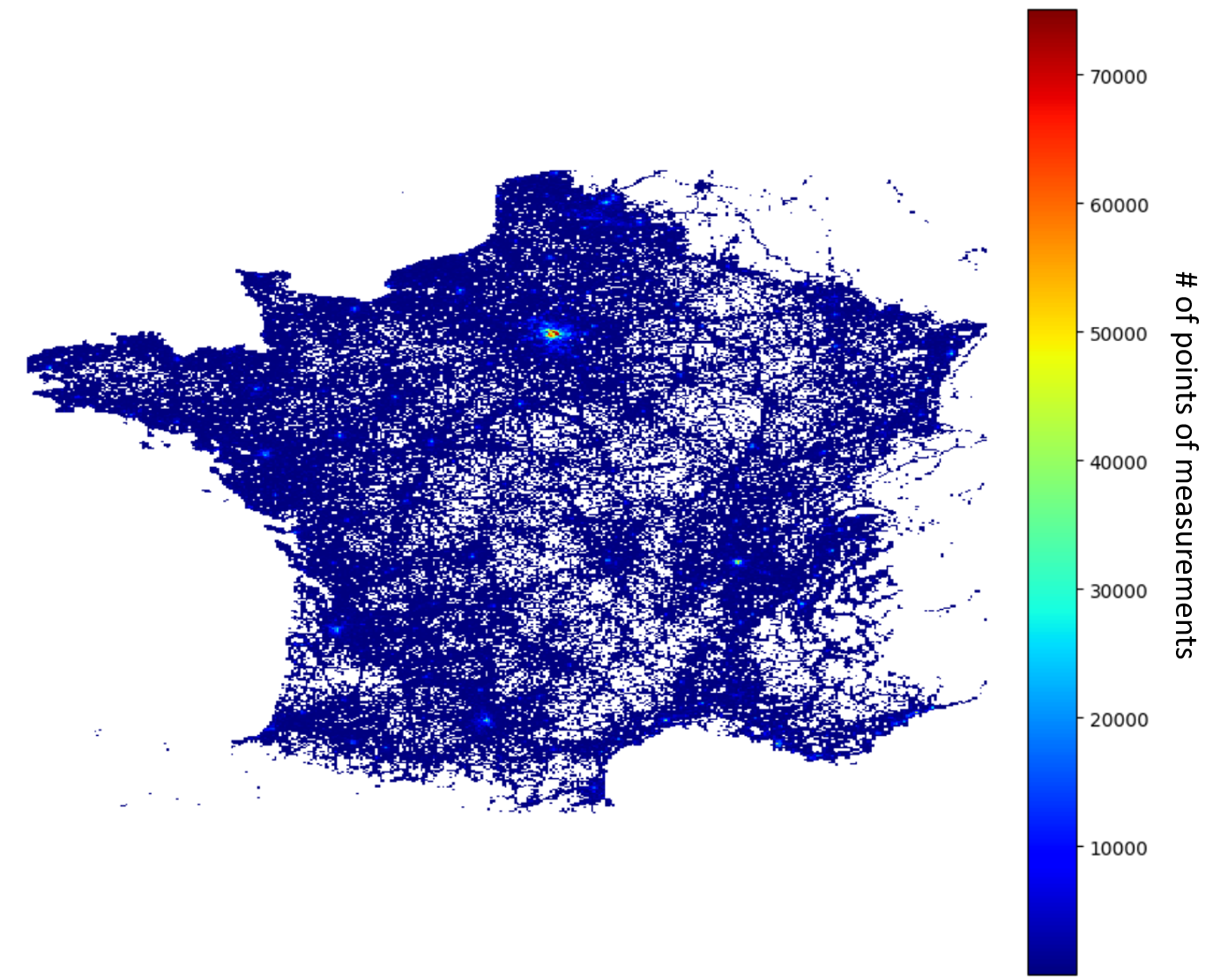}
\caption{2D histogram of the number of measurement points in France.}
\label{francepoint}
\end{figure}

\begin{figure}[!htb]
\includegraphics[scale=0.18]{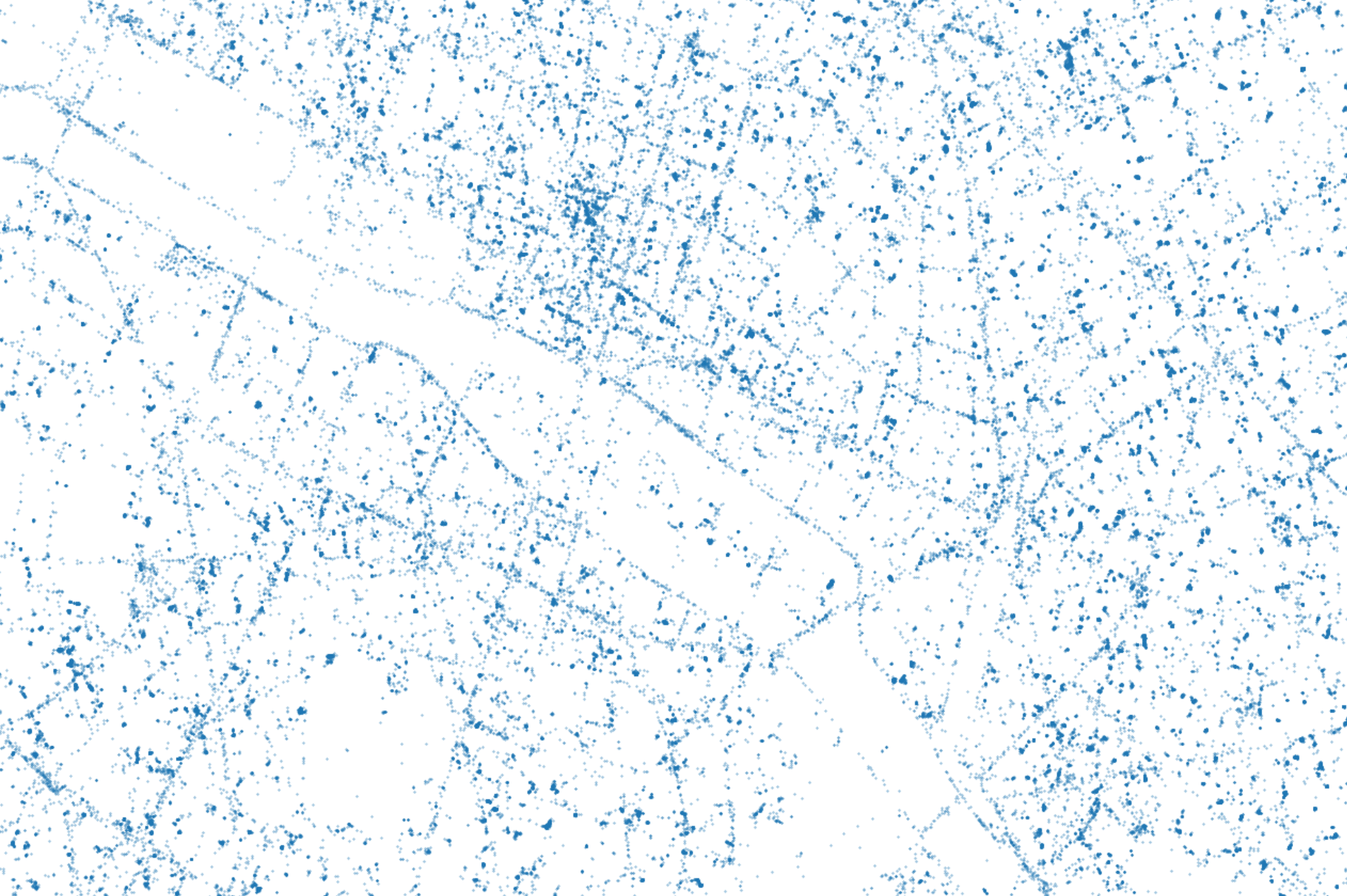}
\caption{A view of all the measurement points in the city of Paris (specificly around Île de la Cité). We can recognize the street structure of the city which indicate the good quality of the GPS information in the dataset. Each blue points is a measurement point.}
\label{paris}
\end{figure}

To generate a comprehensive dataset, we began by creating images of the geographical information surrounding the antennas. We limited our image creation to a 2km x 2km area with a 5-meter resolution, resulting in images of 400 x 400 pixels in size.

\begin{figure}[!htb]
\includegraphics[scale=0.55]{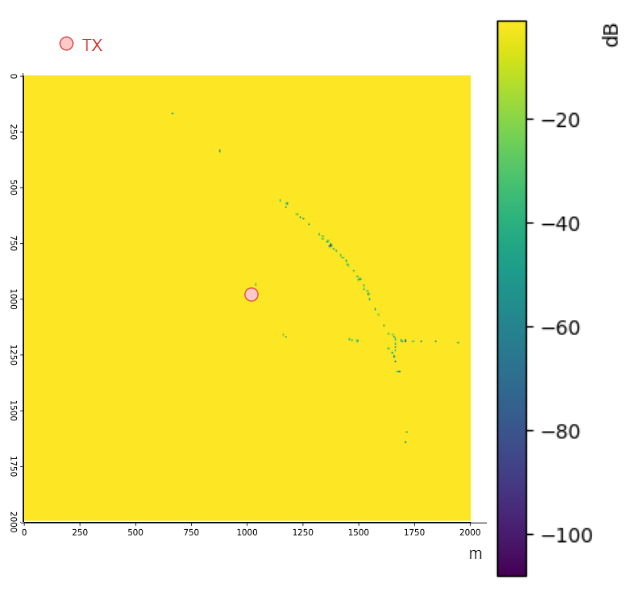}
\caption{This figure provides a visual representation of the measurement points surrounding the antenna. The antenna is located at the center of the figure, and all points that appear in yellow represent areas where no measurements were taken, and thus have a default value of -1. The remaining points in the figure display the RSRP received at the specific location by users, which are influenced by the specific configuration of the antenna diagram.}
\label{exemplemeasure}
\end{figure}

We created images that are center on the cell antenna positions and we add the point of measurement around the antenna to create a picture of measurement around the cell (example in figure \ref{exemplemeasure}).

\subsection{Geographic dataset}

Various types of geographic information around the antenna are relevant for radio propagation modeling, such as the topology of nearby buildings (including their type and height). In this work we limit ourself to a square of 2kmx2km around the antenna located at the center of the image. We take a resolution of 5m. In future work, we plan to incorporate information about the types of materials used in these buildings to gain more insight into how radio waves interact with different materials.

To obtain geographic data, we explored open-source datasets in France, such as those provided by the IGN (Institut national de l'information géographique et forestière / National Institute of Geographic and Forestry Information, https://ign.fr/) or the BNB (base nationale des bâtiments), as well as private datasets. Ultimately, we extracted and rasterized (going from vector data format to image data format) information about all the buildings in the area surrounding the antenna (which is located in the center of the image).

The information about the building configuration is important because allows us to estimate the impact of building on radio propagation and thus improving the prediction power of our model.

\begin{figure}[!htb]
\includegraphics[scale=0.5]{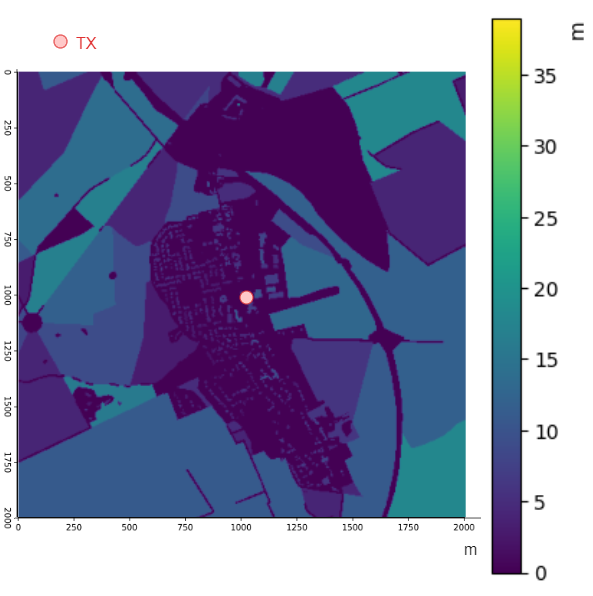}
\caption{A view of the buildings / vegetation heights information around the antenna (at the center of the image). }
\label{heights}
\end{figure}

\begin{figure}[!htb]
\includegraphics[scale=0.5]{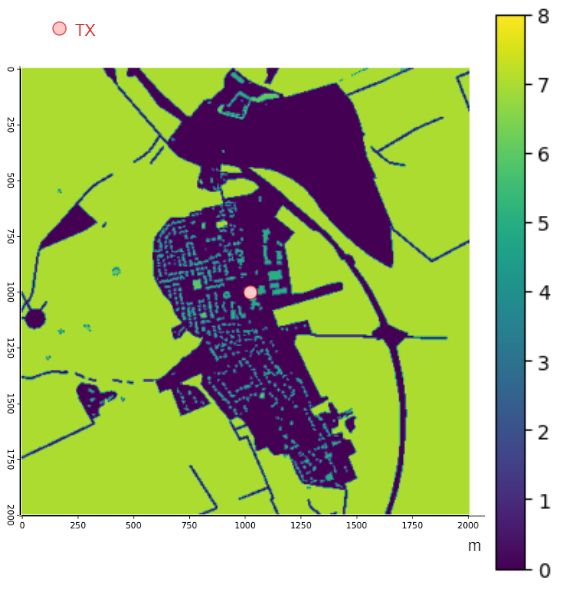}
\caption{A view of the buildings type information around the antenna. There is several classes : 0 is that there is nothing on the ground, 5 is a building, 7 is vegetation and 6 is water / river.}
\label{types}
\end{figure}

\begin{figure}[!htb]
\includegraphics[scale=0.5]{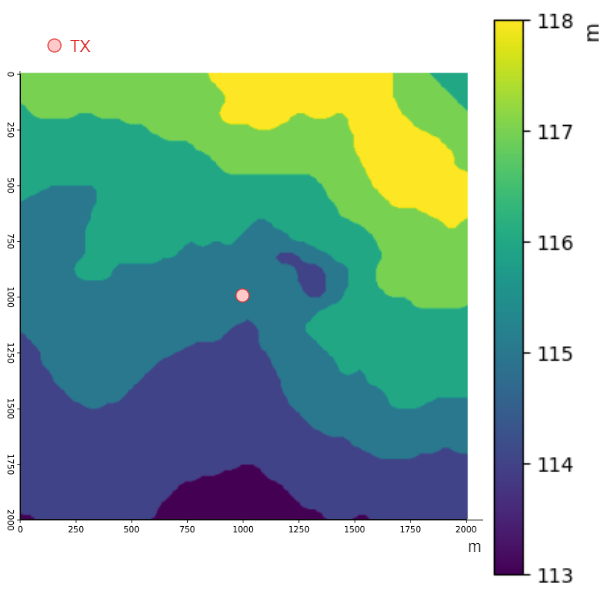}
\caption{A view of the ground height around the antenna}
\label{ground}
\end{figure}

As for the volume of data, we have approximately the geographical information of 20000 sites. We have 3 images representations of the building data : the buildings / vegetation heights (figure \ref{heights}), ground heights (figure \ref{types}), and ground types (figure \ref{ground}).

\subsection{Antennas dataset}

\begin{figure}[!htb]
\includegraphics[scale=0.5]{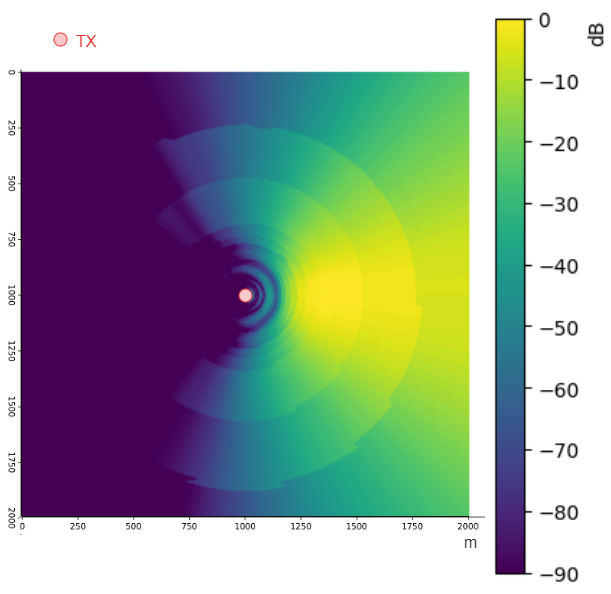}
\caption{The antenna attenuation for a receiver - antenna direct path. The antenna is at the center of the image}
\label{antenna}
\end{figure}

In addition to the geographic dataset, we also incorporate another input into our model to estimate radio propagation around the antenna: the antenna configuration. This includes scalar values such as antenna height, frequency, gain, and EIRP, as well as the antenna diagram. To represent the antenna diagram as input to our model, we generate an image of the antenna diagram attenuation, essentially calculating the attenuation in the direction of the user position (figure \ref{antenna}).

We also use the antenna height ($h_{antenna}$) and frequency ($f$) as inputs to the neural network, which are available for all the antenna cells in the Orange network.

\section{Graph Neural Network Model}

There exists a lot of models that can take image as inputs and output another image-like output (UNET \cite{b12} FNO \cite{b14}). Those models have already been used to model radio propagation \cite{b13}. Interestingly Graph Neural Network can also manage images as input (for example in \cite{b15}) : one just have to convert the image into a graph using the grid like structure of an image as the graph structure. Here the pixels are the nodes and the edges can be represented as links between pixels that are close to each others. 

One of the big advantages of GNN is that they can include invariance properties into the model, thus reducing the need for data. Also one can include implicit physics knowledge into the graph structure : we will exploit those 2 opportunities to improve the model performance.

The global architecture is represented in the appendix but we represent the inputs-outputs in the image below (figure \ref{intout}) :

\begin{figure}[!htb]

\includegraphics[scale=0.6]{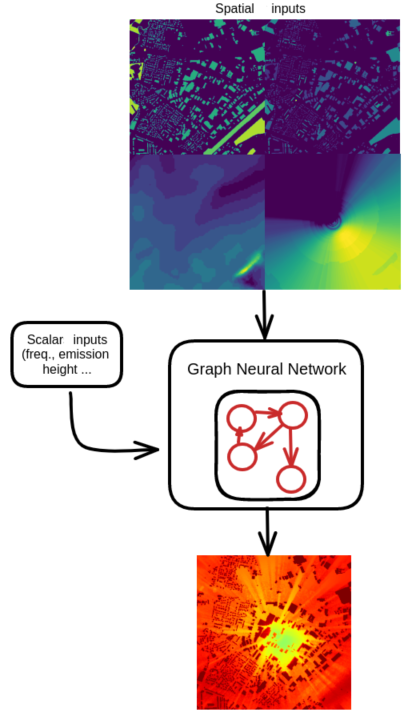}
\caption{The model inputs / outputs}
\label{intout}
\end{figure}

\subsection{Constructing the Input Graph}

We represent the input image as a graph by converting each pixel into a node. We construct two graphs from this node representation to model different factors affecting radio propagation.

\begin{figure}[!htb]
\includegraphics[scale=0.5]{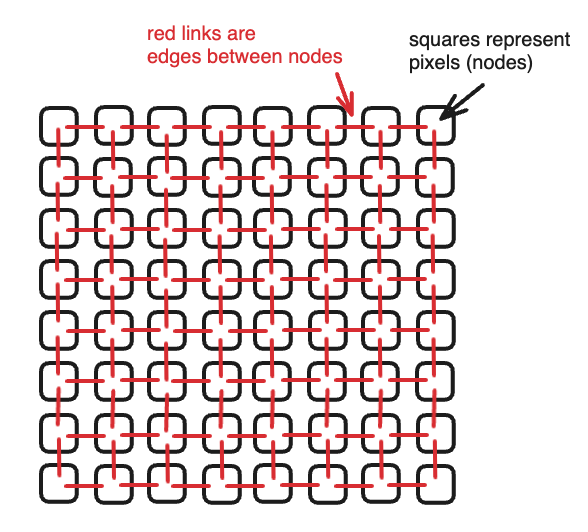}
\caption{Edges for the "grid" graph. This set of edge will enable node to pass information between places that are next to each other.}
\label{graph1}
\end{figure}

\textbf{The first graph} (figure \ref{graph1}) connects nodes that are spatially close in the image, capturing the correlation between received power levels of nearby points due to radio diffusion.

\begin{figure}[!htb]
\includegraphics[scale=0.4]{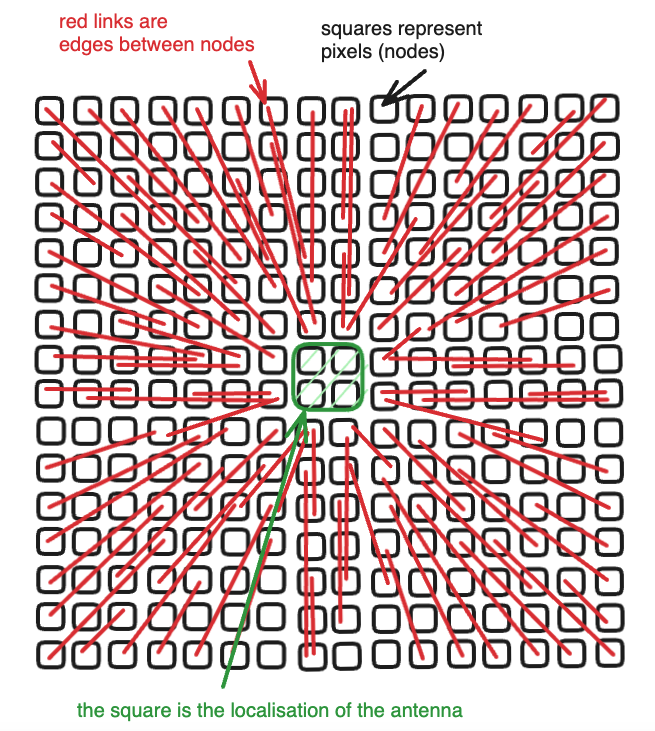}
\caption{Edges for the "ray tracing" graph. This set of edge will enable node to pass information between places that are in alignment with the antenna.}
\label{graph2}
\end{figure}

\textbf{The second graph} (figure \ref{graph2}) connects nodes that are aligned along potential ray tracing paths from the transmitter antenna. This represents the effect of line-of-sight propagation between the antenna and nodes. We link those node without taking in consideration buildings that could mask the line of sight (no visibility condition) propagation so we just have to compute this graph once (regardless of the buildings / ground height structure).

We then pass these two graphs through a graph neural network to learn a model that combines both the influence of spatial node relationships and antenna-node ray tracing on the predicted radio propagation characteristics.
By using two graphs to represent different aspects of the environment, the graph neural network can learn how both diffusion and line-of-sight propagation impact received signal strength.

One can visualize those 2 graphs in the figures \ref{graph1} and \ref{graph2}.

We expect the second graph capturing ray tracing relationships to significantly influence model performance and coverage map predictions. 

Also by representing edges in polar coordinates relative to the antenna, the graph exhibits rotation invariance. In other words, rotating the input graph leads to a corresponding rotation in the predicted radio propagation maps. This invariance is desirable as the ray tracing relationships should be independent of absolute orientation. By introducing this structural property into our graph neural network through the use of polar coordinates, we can better capture the characteristics of radio propagation.

In summary, our graph neural network takes as input a graph with node features of dimension $N_{nodes}\times d_{node}$, where $d_{node}$ is the number of features per pixel and $N_{nodes}$ the number of pixels in the original image. 
The graph contains two sets of edges: grid edges capturing spatial relationships and ray tracing edges capturing ray tracing relationships. 
The grid edges have dimension $N_{edge \,grid}\times2$ and the ray tracing edges have dimension $N_{edge \,ray}\times2$ ($N_x$ being the number of edges and the $2$ columns being the receivers id nodes and the senders id nodes). The edges attributes are 2D polar coordinates $(r, \theta)$ encoding differences in distance and angle $(\Delta r, \Delta \theta)$  to the antenna between connected nodes. 
By passing this multi-graph input through the graph neural network, we can learn how the diverse relationships and features influence radio propagation characteristics.

The graph needs to be compute once at the GNN initialization and will not need to be recompute at each inference.

\subsection{Explanation of the graph neural network architecture}

As our main neural architecture we choose to use the graph network block paradigm \cite{b4} \cite{b6}. 
The final GNN (graph neural network) architecture looks like the figure 12. It is composed of several elements :

- \textbf{the node encoder and the edge encoder} : those are simply a MLP (Multi Layer Perceptron) that project the nodes of dimension $(N_{nodes}, d_{nodes})$ ($N_{nodes}$ being the number of nodes in the graph and $d_{nodes}$ the original dimension of the input node, here 4 for the 4 spatials inputs) and the edges of dimension $(N_{edges}, d_{edges})$ ($N_{edges}$ being the number of edges in the graph and $d_{edges}$ the original dimension of the input edges, here 2), to standard dimension $(N_{nodes}, d_{encoder})$ and $(N_{edges}, d_{encoder})$. $d_{encoder}$ is the output dimension of the encoder.

- \textbf{the FiLM layers} \cite{b17} (Feature-wise Linear Modulation) is used to combine scalar inputs with spatial inputs. It is inspired by \cite{b18}. The FiLM layer is typically used in conditional deep learning tasks : here we used it to condition the radio propagation to scalar inputs that are the frequency of emision, the antenna height and the EIPR. It's a simple transformation of the nodes features $x$ : $\text{FiLM}(x) = \gamma \odot x + \beta$ where $\gamma$ and $\beta$ are values computed with another MLP.

- \textbf{the GraphNetwork block} \cite{b6} : it is used to propagate messages though the graph (and so nodes features can influence each other in order to improve performance). There is multiple message-passing round with different weights each time. We did test multiple types of GNN (GAT \cite{b27}, graphnetwork block \cite{b6}) and we obtain similar results.

- \textbf{the node decoder} : a MLP (Multi Layer Perceptron) that map the nodes dimension $(N_{nodes}, d_{decoder})$ to $(N_{nodes}, d_{output})$ where $d_{output})$.

\subsection{Details on the training process}

In this section, we provide details on the training process for our data-driven radio propagation model. We utilized PyTorch and PyTorch Geometric to code the graph neural network and trained the model with various parameters such as learning rate, number of graph network blocks, and dimension of the encoder/decoder output. The training was performed on an NVIDIA GPU A100 40G and took approximately 100 hours to complete.

To train the graph neural network, we adopted a semi-supervised approach that allows us to train the GNN with partial output targets, as described in \cite{b16}. Specifically, we employed the masked output training procedure explained in Section II.C to train the GNN with partial target values.

As our problem is a regression task, we utilized the $L_2$ loss function between the predicted attenuation ($\hat{p}_i$) and the actual measurement ($p_i$). To update the neural network, we computed the resulting propagation map with the GNN ($f(x_{spatial}, x_{scalar})$) and then calculated the error loss on the points where we have actual measurement values ($L_{map_i} = \sum_{j \in M} (f(x_{spatial}, x_{scalar})[pos_j] - p_{j})^{2}$).
In the above equation, $pos_j$ represents the spatial position of the $j$-th measurement point and $M$ represents the set of all measurement points where we have actual measurement values.


\subsection{Evaluation metrics used to assess model performance}

The main evaluation metric we use is the root mean square error (RMSE), which measures the difference between the predicted signal strength values and the actual values from the validation dataset.

One of the key point to make is that we clearly separate the training set from the validation set by separating sites between the training set and the validation set (so one radio sites cannot be in both the validation set and the training set). This is done to avoid data leakage and performance overestimation \cite{b21}.

\section{Results and Analysis}

In this section we will details the training procedure and make comparisons with different type of models (physics-based models  / heuristics / and different types of GNN).

\subsection{Training convergence and performance}

The training was done using the Adam optimizer \cite{b24} with a learning rate of $1. 10^{-4}$. Due to the high memory requirement of GNN, we could only train one image (graph) per batch. So we technically have a batch size of only one.
Here we plot the training loss evolution in the annex section. Three of notable elements during the training :

- The model doesn't overfit the train set (the loss didn't go to 0 for the training loss) as it is often the case with deep learning model. It is possible that the dataset have a lot of noise and the model can't forecast this pure noise.

- The training is unstable : we don't have a smooth convergence toward a lower training loss value. We could have stabilized the train loss variation by improving the batch size but the memory requierements to do that were too high (our GPU didn't have enough memory to increase the batch size).

- There is a very fast convergence as it seems that after 50k training steps the model has already converge.

Below a comparison of performance between differents models :

\begin{table}[ht]
\centering
\caption{Comparison of performance on the raw dataset}
\label{table:algorithm_comparison}
\begin{tabular}{@{}lcccc@{}}
\toprule
\textbf{Algorithm} & \textbf{RMSE (test)} \\ \midrule
GNN with ray tracing      & 9.8dB      \\
GNN without ray tracing   & 10.5dB     \\
Tabular model             & 10.2dB     \\
\bottomrule
\end{tabular}
\label{tab:table1}
\end{table}

Our model's forecasting capabilities were improved by incorporating a specially designed graph neural architecture that is capable of handling the ray-like behavior observed in radio propagation. It is a classic approach in neural network design to introduce structural biases that are specific to the problem domain, and doing so has been shown to improve the accuracy and robustness of predictions in various applications. The tabular model corresponds to classic tabular model (gradient boosting) that use only the nodes features inputs to make prediction on the same node / pixel (it makes a prediction without taking into account the surrounding geographical environment).

We also conduct other experiments focusing on outdoor prediction accuracy : we filter extrems values of the dataset (train and test) (notably we remove data measurements that are below -110dB that are indoor). Also we filter the data measurements on agent that have a speed $>$ 10km/h : it assures that the agents is moving at speed that implies they are not indoor where it is difficult for the model to make an accurate prediction. Below the comparaison :

\begin{table}[ht]
\centering
\caption{Comparison of performance on the raw dataset with indoor filter}
\label{table:algorithm_comparison}
\begin{tabular}{@{}lcccc@{}}
\toprule
\textbf{Algorithm} & \textbf{RMSE (test)} \\ \midrule
GNN with ray tracing      & 8.5dB         \\
GNN without ray tracing   & 8.9dB         \\
Tabular model             & 9.1dB       \\
\end{tabular}
\label{tab:table2}
\end{table}

On purely outdoor dataset the model is competitive with the custom physical model that is a legacy model from Orange. Nevertheless the physical model takes a lot more time to compute the radio coverage map ($\approx$ 5s vs $<$0.500s for the GNN with GPU).

\begin{table}[ht]
\centering
\caption{Comparison of performance in term of speed of execution}
\label{table:algorithm_comparison}
\begin{tabular}{@{}lcccc@{}}
\toprule
\textbf{Algorithm} & \textbf{Speed} \\ \midrule
GNN with ray tracing      & 0.18s (GPU)   \\
GNN without ray tracing   & 0.14s (GPU)   \\
Tabular model             & 0.035s (CPU)        \\
Physical model            & $\approx$ 5s  (CPU)  \\ \bottomrule
\end{tabular}
\label{tab:table3}
\end{table}

Some precision about the "speed of execution" metric : it is the speed to compute the 400x400 pixels propagation map (2kmx2km at 5m resolution).

\subsection{Discussion of the key factors that affect radio propagation, as identified by the model}

\begin{figure}[!htb]
\includegraphics[scale=0.48]{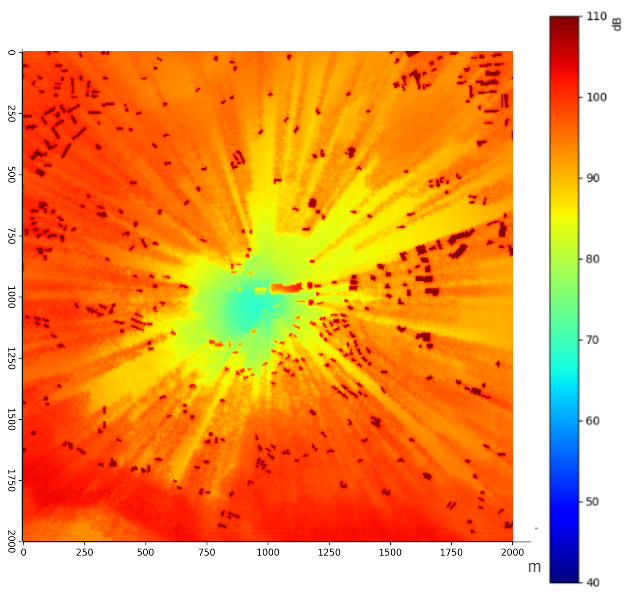}
\caption{Visual of the propagation map estimated with ray tracing capabilities (taking into account ray-like edges into the input graph)}
\label{rayimage}
\end{figure}

\begin{figure}[!htb]
\includegraphics[scale=0.48]{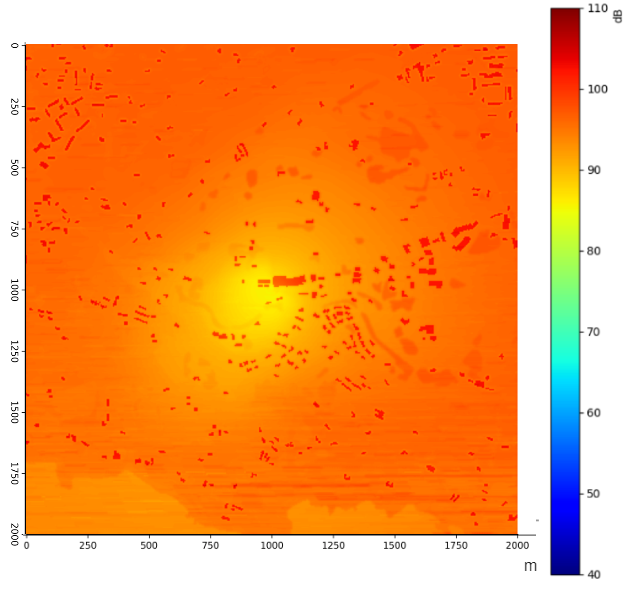}
\caption{Visual of the propagation map estimated without ray tracing capabilities (not taking into account ray-like edges into the input graph)}
\label{norayimage}
\end{figure}

One can see the impact of the ray tracing edge on the final result in term of quality of predictions. Also the final visualisation is also very different in term final result (figure \ref{rayimage} vs figure \ref{norayimage}).
Here an example of the propagation result with ray tracing edges and without raytracing edges.

The model (the ray tracing one) is capable of accurately capturing the impact of obstacles located between the antenna and the receiver, which results in a reduction of the received power.

By purely visual interpretation of the coverages map, the model also accounts for the significant reduction in received power levels inside buildings compared to outdoor locations. Additionally, the model estimates higher received power levels in the direction of the antenna orientation, by considering the antenna direction/diagram as a contributing factor (appendix section to observe more coverage map estimation).

\section{Future Work}

\subsection{Areas for future research}

One promising direction for future work is exploring modifications to the graph neural network architecture. The particular GNN architecture used in this work could be adjusted by changing the number of layers and hidden units, or using different message passing approaches such as graphformer \cite{b22}. These architectural changes may allow the GNN to better capture the complex relationships in radio propagation. For example, deeper layers or residual connections could enable learning longer-range dependencies, while different message passing methods may be more suited to modeling the spatial and ray tracing relationships in the graph. Testing alternative architectures could reveal insights into the strengths and limitations of GNNs for this application and lead to accuracy improvements.

Another promising direction for future work is incorporating physical loss characteristics into the graph neural network like the Physic-Informed Neural Network loss \cite{b23} (PINN loss). Sources of loss such as free space loss, material absorption, and diffraction can be encoded as edge or node attributes in the input graph. The GNN can then learn to integrate knowledge of these physical loss mechanisms into its propagation predictions. 

One potential area of future work is to improve the machine learning methodology. In our approach we try to directly minimize the $L2$ error in order to calibrate the weights of the GNN. But one other approach and perhaps more accurate one would be to maximize the likelihood of the radio mapping using something like conditional GAN \cite{b25}. We let this avenue for future work.

\subsection{Limitations of the current study}

One of the main limitations of our study is that it is based on data point measurements, which can introduce bias into the model. Specifically, the bias arises from the fact that we only get measurements at locations where we are able to perform the measurements, which typically excludes areas behind the antenna where users do not receive a direct signal. Instead, data points measurements in these areas are typically the result of reflections off buildings, which can lead to an overestimation of the radio power received behind the antenna (the model only see data points measurement that are the results of reflections behind antennas and wrongly overestimate the radio power received behind antennas).

The bias in our model due to the limited availability of data points is an example of survivor-ship bias \cite{b26}, which occurs when we only consider the data that has survived a particular selection process (here the user have selected the cell with the strongest received power).

\section{Conclusion}

In conclusion, our research has demonstrated the effectiveness of using graph neural networks for data-driven radio propagation modeling. By leveraging the power of machine learning, we were able to construct a model that accurately predicts radio power levels in complex urban environments, taking into account the effects of buildings, terrain, and other obstacles on radio wave propagation.

Our approach provides a promising framework for future research in this area, offering a data-driven alternative to traditional models that rely on complex mathematical calculations and approximations. By training our model on a comprehensive dataset of geographical information and radio power measurements, we were able to achieve high levels of accuracy and improve upon existing models that are limited in their ability to account for the complexities of real-world environments.

\section{Appendix} %

This section contains details regarding the model architecture and training, with a focus on providing precise information about the model parameters. The aim is to enable accurate reproduction of the neural network architecture.

\subsection{Model global architecture} %

\begin{figure}[!htb]
\includegraphics[scale=0.25]{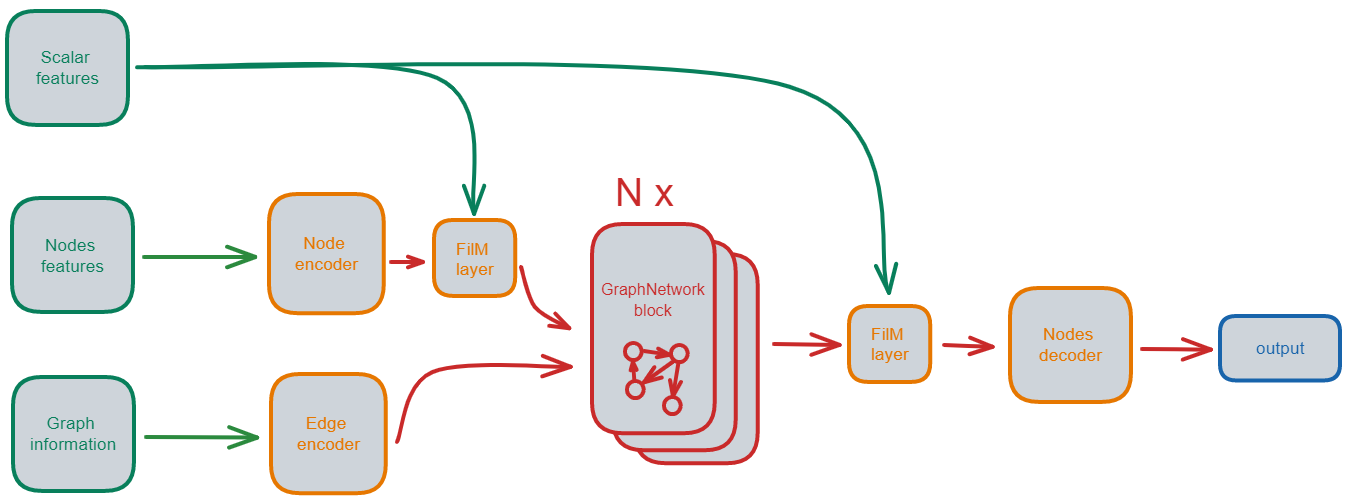}
\caption{The model architecture and different components}
\label{archi}
\end{figure}

The overall architecture of the neural network (figure \ref{archi}) is a standard encoder-preprocess-decoder graph neural network, which also takes inputs from scalar features. The input features are color-coded in \textcolor{green!50!black}{green}, with the scalar features represented as a concatenation vector of cell frequency, antenna height, and EIPR, and the node features as spatial information corresponding to a concatenation of building type information, building height, ground height, and antenna diagram loss in LoS (line of sight). The graph information contains all the information at the edge level, including the relationship between the different pixels/nodes of the graph, represented here by the difference in polar coordinates between the pixels/nodes ($\Delta \theta$, $\Delta r$).

The various components of the graph neural network are color-coded in \textcolor{red}{red} and \textcolor{orange}{orange}. The different neural network components are already detailed in section IV.B. Finally, the output propagation map is represented in \textcolor{blue}{blue}.

\newpage

Here a more complet view of the different neural network parameters :

\begin{table}[ht]
\caption{Model Parameters}
\centering
\begin{tabular}{p{2cm} l p{5cm}}
\hline\hline
Parameter & Value & Description \\
\hline
Learning rate & 0.0001 & step size for gradient descent \\
Batch size & 1 & number of samples in each mini-batch \\
Nb of epochs & 10 & number of iterations over the training data \\
Hidden layer size & 128 & number of neurons in each hidden layer \\
Nb of hidden layers (encoders) & 2 & number of hidden layers in the encoders \\
Nb of hidden layers (decoders) & 2 & number of hidden layers in the decoders \\
Nb of message passing blocks (N) & 10 & number of message passing layers (GNN layers)\\
Activation function & ReLU & activation function used in each neuron \\
Optimizer & Adam & optimization algorithm used for gradient descent \\
Loss function & MSE & objective function optimized by the model \\
Total number of parameters & $1.6*10^6$ & number of parameters the model has to learn \\
\hline
\end{tabular}
\label{tab:model-parameters}
\end{table}

\subsection{Training loss evolution} %

Below the loss curve of a training session :

\begin{figure}[!htb]
\includegraphics[scale=0.2]{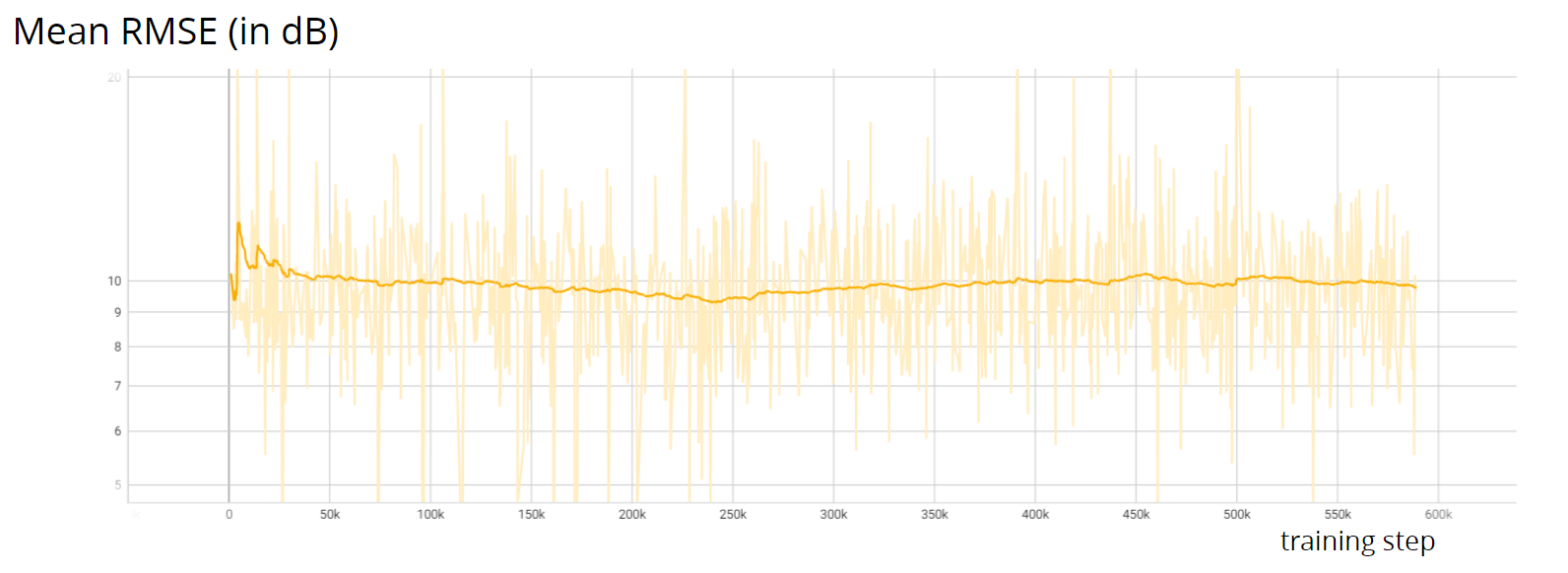}
\caption{Evolution of the loss value during training. One can see that the loss rapidly stabilize around 9-10dB}
\end{figure}

\subsection{Map coverage visualisations} %

\begin{figure}[!htb]
\includegraphics[scale=0.44]{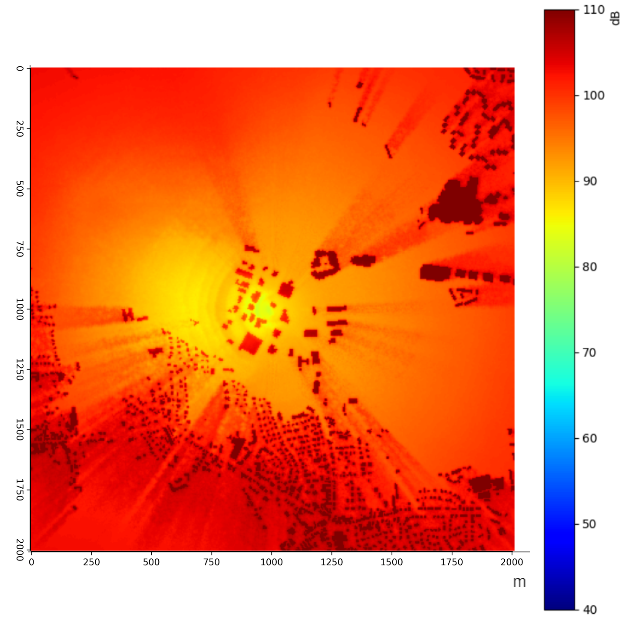}
\caption{Example of a coverage map for frequency 2600Mhz}
\label{propag1}
\end{figure}

\begin{figure}[!htb]
\includegraphics[scale=0.44]{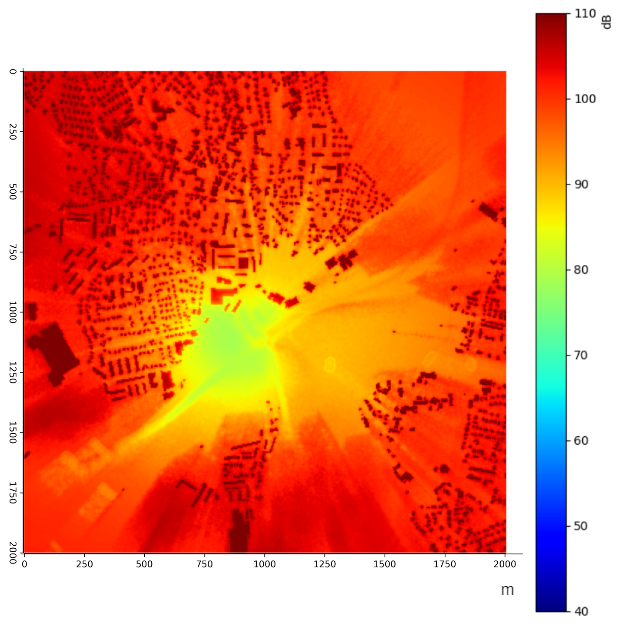}
\caption{Example of a coverage map for frequency 2100Mhz }
\label{propag2}
\end{figure}

\begin{figure}[!htb]
\includegraphics[scale=0.44]{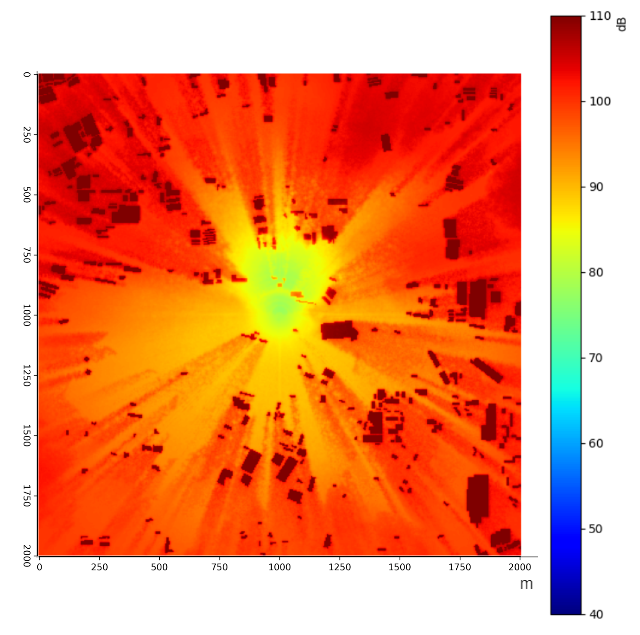}
\caption{Example of a coverage map for frequency 2100Mhz }
\label{propag3}
\end{figure}

We provide a series of examples of what the model outputs as full radio map coverage (figures \ref{propag1} \ref{propag2} \ref{propag3}).

\label{appendix} 

\section*{Acknowledgment}

We acknowledge that portions of the paper text, including the abstract, problem statement, novelty section and references, were generated with the assistance of AI-based tools. However, the identification of the research problem, choice of approach, experiment design, data analysis and interpretation of the results remain the intellectual work of the authors.

The success of this research depended crucially on the availability of wireless network data. We thank the network operators who provided access to these datasets.

\vspace{12pt}

\end{document}